\pdfoutput=1

\documentclass[11pt]{article}

\usepackage[]{EMNLP2022}

\usepackage{times}
\usepackage{latexsym}

\usepackage[T1]{fontenc}

\usepackage[utf8]{inputenc}

\usepackage{microtype}

\usepackage{inconsolata}

\usepackage{times}
\usepackage{latexsym}
\usepackage{graphics}
\usepackage{graphicx}
\usepackage{color}
\usepackage{colortbl}
\usepackage{amssymb}
\usepackage{amsmath}
\usepackage{bbm}
\usepackage{booktabs}
\usepackage{microtype}
\usepackage{multirow}
\usepackage{footmisc}
\usepackage{soul}
\usepackage{arydshln}

\usepackage{algpseudocode}
\usepackage{algorithm}
\usepackage{hyperref}
\usepackage{xspace}

\newcommand{\yy}{\mathbf{y}}

\newcommand{\calY}{\mathcal{Y}}

\newcommand{\vtheta}{{\boldsymbol \theta}}

\newcommand{\vocab}{\mathcal{V}}

\newcommand{\model}{q}

\definecolor{darkblue}{rgb}{0.0,0.0,0.5}
\definecolor{purple}{rgb}{0.5,0.0,0.5}

\newcommand{\eg}{\textit{e.g.}}

\usepackage{color}
\usepackage{tikz}
\usepackage[edges]{forest}
\definecolor{hiddendraw}{RGB}{205, 44, 36}
\definecolor{hidden-blue}{RGB}{194,232,247}
\definecolor{hidden-orange}{RGB}{243,202,120}
\definecolor{hidden-yellow}{RGB}{242,244,193}
\tikzstyle{mybox}=[
    rectangle,
    draw=hiddendraw,
    rounded corners,
    text opacity=1,
    minimum height=1.5em,
    minimum width=5em,
    inner sep=2pt,
    align=center,
    fill opacity=.5,
    ]

\usepackage{amssymb}
\usepackage{pifont}

\usepackage{amsthm}

\usepackage{multirow} 

\usepackage{subfigure}
\usepackage{mdframed}
\DeclareCaptionFont{black}{\color{black}}
\newcommand{\foo}{\color{cyan}\makebox[0pt]{\textbullet}\hskip-0.5pt\vrule width 1pt\hspace{\labelsep}}

\title{Generative Knowledge Graph Construction: A Review}

\author{
Hongbin Ye\textsuperscript{\rm 1,2}, 
Ningyu Zhang\textsuperscript{\rm 1,2 \thanks{\quad Corresponding author.}},
Hui Chen \textsuperscript{\rm 3},
Huajun Chen\textsuperscript{\rm 1,2}  \\
\textsuperscript{\rm 1} Zhejiang University \& AZFT Joint Lab for Knowledge Engine\\
\textsuperscript{\rm 2} Hangzhou Innovation Center, Zhejiang University \\
\textsuperscript{\rm 3}  Alibaba Group  \\
\texttt{\{yehongbin,zhangningyu,huajunsir\}@zju.edu.cn},weidu.ch@alibaba-inc.com \\
}

\begin{document}
\maketitle
\begin{abstract}

Generative Knowledge Graph Construction (KGC) refers to those methods that leverage the sequence-to-sequence framework for building knowledge graphs, which is flexible and can be adapted to widespread tasks. In this study, we summarize the recent compelling progress in generative knowledge graph construction. We present the advantages and weaknesses of each paradigm in terms of different generation targets and provide theoretical insight and empirical analysis. Based on the review, we suggest promising research directions for the future. Our contributions are threefold: (1) We present a detailed, complete taxonomy for the generative KGC methods; (2) We provide a theoretical and empirical analysis of the generative KGC methods; (3) We propose several research directions that can be developed in the future.

\end{abstract}
\section{Introduction}

Knowledge Graphs (KGs) as a form of structured knowledge have drawn significant attention from academia and the industry \cite{DBLP:journals/tnn/JiPCMY22}. 
However, high-quality KGs rely almost exclusively on human-curated structured or semi-structured data. 
To this end, Knowledge Graph Construction (KGC) is proposed, which is the process of populating (or building from scratch) a KG  with new knowledge elements (e.g., entities, relations, events). 
Conventionally, KGC is solved by employing task-specific discriminators for the various types of information in a pipeline manner \cite{DBLP:conf/tac/AngeliZCCBPPGM15,DBLP:conf/emnlp/LuanHOH18,DBLP:conf/emnlp/MesquitaCSMB19,zhang2022deepke}, typically including 
(1) entity discovery or named entity recognition \cite{DBLP:conf/conll/SangM03}, 
(2) entity linking \cite{DBLP:conf/cikm/MilneW08},
(3) relation extraction \cite{DBLP:journals/jmlr/ZelenkoAR03}
and (4) event extraction \cite{DBLP:conf/emnlp/DuC20}.
However, this presents limitations of error population and poor adaptability for different tasks. 

\begin{figure}
    \centering
    \resizebox{.48\textwidth}{!}{
    \includegraphics{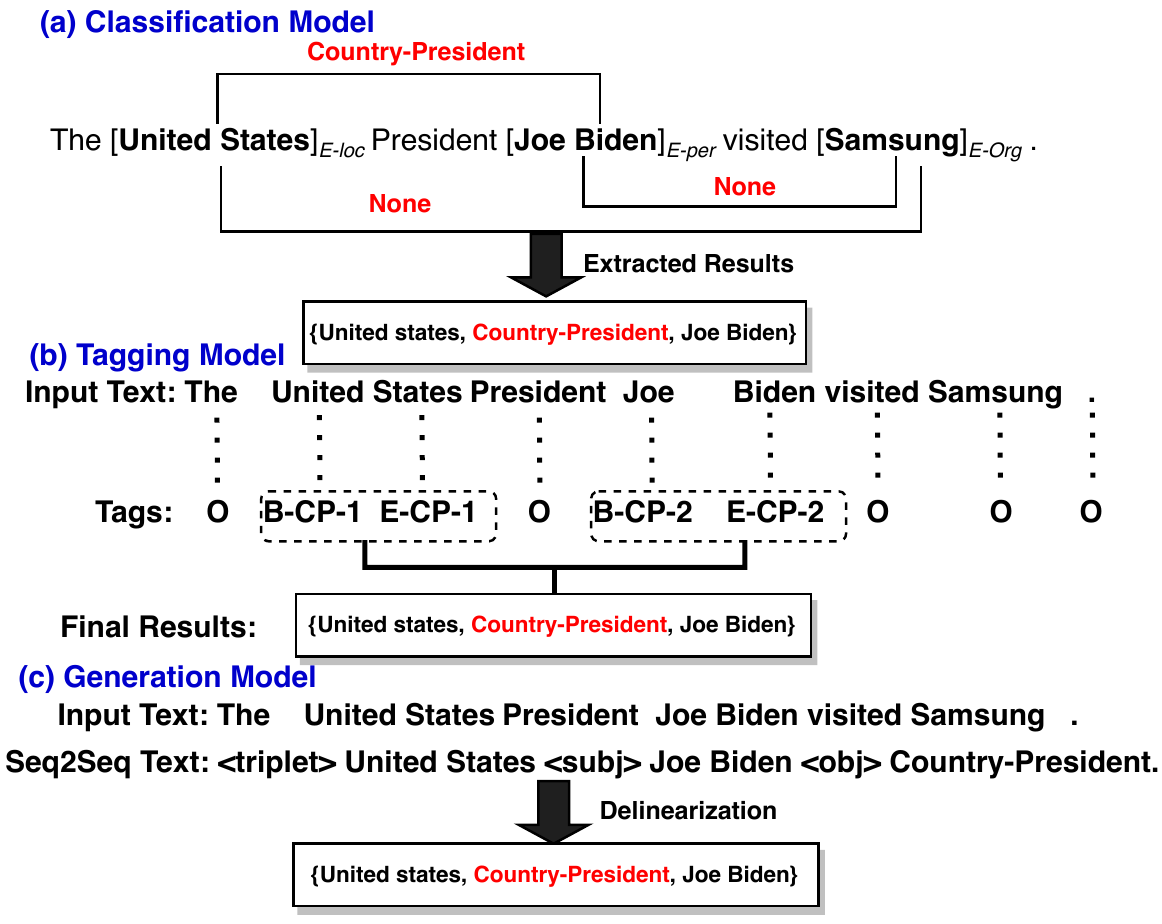}}
    \caption{
    Discrimination and generation methodologies for relation extraction.
    “Country-President” is the relation, and “CP” is short for “Country-President.”}
    \label{fig:intro}

\end{figure}

\paragraph{Generative Knowledge Graph Construction.} 
Some generative KGC methods based on the sequence-to-sequence (Seq2Seq) framework are proposed to overcome this barrier.
Early work \cite{DBLP:conf/acl/LiuZZHZ18} has explored using the generative paradigm to solve different entity and relation extraction tasks. 
Powered by fast advances of generative pre-training such as T5 \citep{DBLP:journals/jmlr/RaffelSRLNMZLL20}, and BART \citep{DBLP:conf/acl/LewisLGGMLSZ20}, Seq2Seq paradigm has shown its great potential in unifying widespread NLP tasks.
Hence, more generative KGC works \cite{DBLP:conf/acl/YanDJQ020,DBLP:conf/iclr/PaoliniAKMAASXS21,DBLP:conf/acl/0001LDXLHSW22} have been proposed, showing appealing performance in benchmark datasets.
Figure~\ref{fig:intro} illustrates an example of generative KGC for relation extraction.
The target triple is preceded by the tag <triple>, and the head entity, tail entity, and relations are also specially tagged, allowing the structural knowledge (corresponding to the output) to be obtained by inverse linearization.
Despite the success of numerous generative KGC approaches, these works scattered among various tasks have not been systematically reviewed and analyzed.

\paragraph{Present work} 
In this paper, we summarize recent progress in generative KGC (An timeline of generative KGC can be found in Appendix \ref{sec:timeline})
and maintain a public repository for research convenience\footnote{\url{https://github.com/zjunlp/Generative_KG_Construction_Papers}}. 
We propose to organize relevant work by the generation target of models and also present the axis of the task level (Figure \ref{taxonomy_of_generation}):

\begin{itemize}
    \item \textbf{Comprehensive review with new taxonomies}. 
    We conduct the \textbf{first} comprehensive review of generative KGC together with new taxonomies.
    We review the research with different generation targets for KGC with a comprehensive comparison and summary (\S \ref{tax}).
    \item \textbf{Theoretical insight and empirical analysis}. 
    We provide in-depth theoretical and empirical analysis for typical generative KGC methods, illustrating the advantages and disadvantageous of different methodologies as well as remaining issues (\S \ref{analysis}). 
    \item \textbf{Wide coverage on emerging advances and outlook on future directions}. 
    We provide comprehensive coverage of emerging areas, including prompt-based learning.  
This review provides a summary of generative KGC and highlights future research directions (\S \ref{future work}). 
\end{itemize}

\paragraph{Related work}

As this topic is relatively nascent, only a few surveys exist.
Closest to our work, \citet{DBLP:journals/tnn/JiPCMY22} covers methods for knowledge graph construction, representation learning, and applications, which mainly focus on general methods for KGC.
\citet{DBLP:journals/corr/abs-2202-05786} provides a systematic survey for multi-modal knowledge graph construction and review the challenges, progress, and opportunities. 
For general NLP, \citet{DBLP:journals/corr/abs-2111-01243} survey recent work that uses these large language models to solve tasks via text generation approaches, which has overlaps in generation methodologies for information extraction.
Different from those surveys, in this paper, we conduct a literature review on generative KGC, hoping to systematically understand the methodologies, compare different methods and inspire new ideas.

\begin{figure}
    \centering
    \resizebox{.48\textwidth}{!}{
    \includegraphics{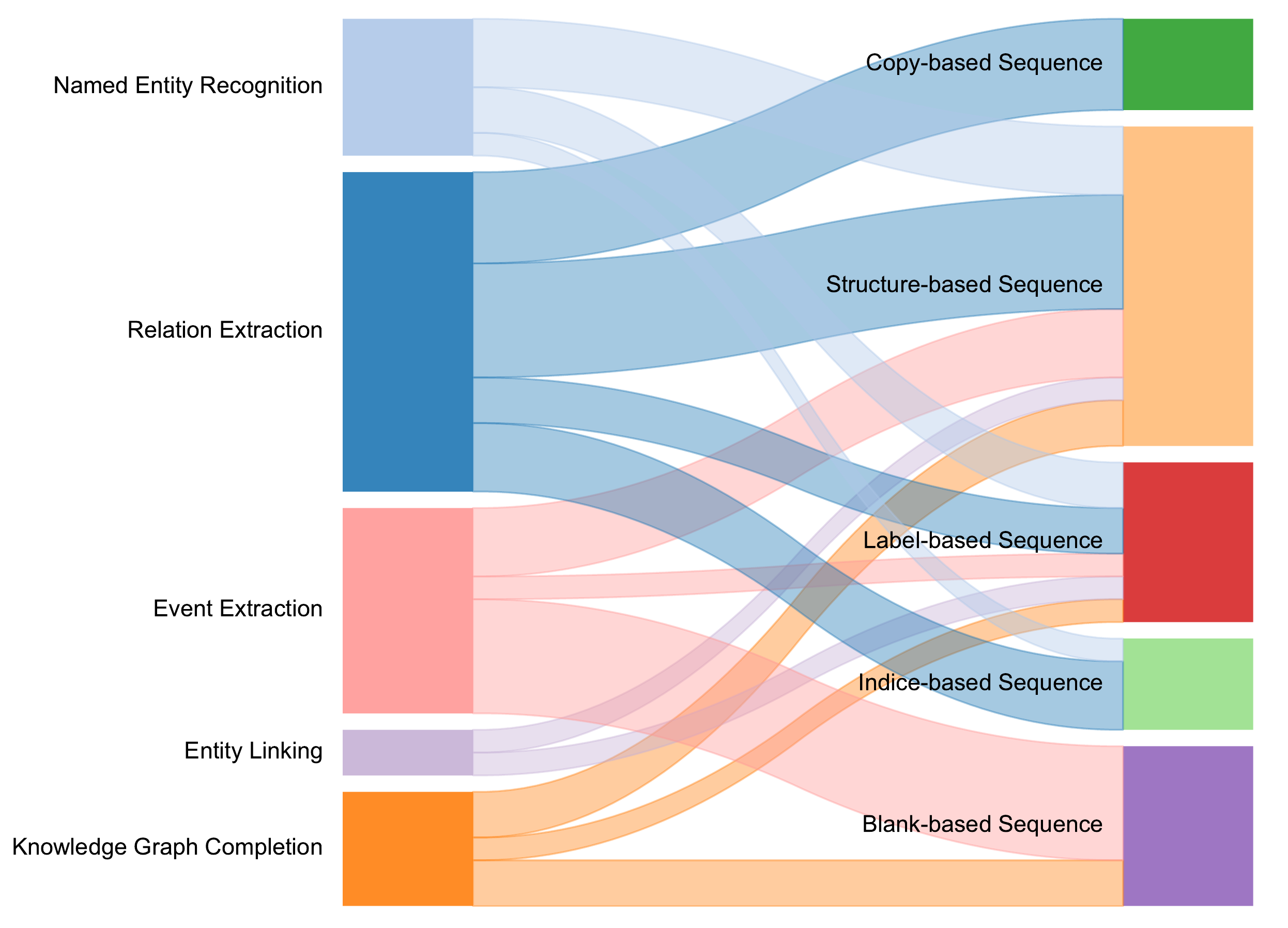}}
    \caption{Sankey diagram of knowledge graph construction tasks with different generative paradigms.}
    \label{fig:tongji}
\end{figure}

\section{Preliminary on Knowledge Graph Construction}

\subsection{Knowledge Graph Construction}

Knowledge Graph Construction mainly aims to extract structural information from unstructured texts, 
such as Named Entity Recognition (NER) \citep{TACL2016_NER_BiLSTM}, Relation Extraction (RE) \citep{EMNLP2015_RE_PCNN}, Event Extraction (EE) \citep{ACL2015_EE_DMCNN}, Entity Linking (EL) \citep{DBLP:journals/tkde/ShenWH15}, and Knowledge Graph Completion \citep{DBLP:conf/aaai/LinLSLZ15}.

Generally, KGC can be regarded as structure prediction tasks, where a model is trained to approximate a target function $F(x) \rightarrow y$, where $x \in \mathcal{X}$ denotes the input data and $y \in \mathcal{Y}$ denotes the output structure sequence. 
For instance, given a sentence, "\emph{Steve Jobs and Steve Wozniak co-founded Apple in 1977.}": 

\noindent
\textbf{Named Entity Recognition} aims to identify the types of entities, \eg, ‘\emph{Steve Job}', ‘\emph{Steve Wozniak}' $\Rightarrow$ \texttt{PERSON}, ‘\emph{Apple}' $\Rightarrow$ \texttt{ORG}; 

\noindent
\textbf{Relation Extraction} aims to identify the relationship of the given entity pair $\langle$\emph{Steve Job}, \emph{Apple}$\rangle$ as \texttt{founder};

\noindent
\textbf{Event Extraction} aims to identify the event type as \texttt{Business Start-Org} where ‘\emph{co-founded}' triggers the event and (\emph{Steve Jobs}, \emph{Steve Wozniak}) are participants in the event as \texttt{AGENT} and \texttt{Apple} as \texttt{ORG} respectively.

\noindent
\textbf{Entity Linking} aims to link the mention \emph{Steve Job} to \texttt{Steven Jobs (Q19837)} on Wikidata, and \emph{Apple} to \texttt{Apple (Q312)} as well.

\noindent
\textbf{Knowledge Graph Completion} aims to complete incomplete triples $\langle$\emph{Steve Job}, \emph{create},  \emph{?}$\rangle$ for blank entities \texttt{Apple}, \texttt{NeXT Inc.} and \texttt{Pixar}.

\tikzstyle{leaf}=[mybox,minimum height=1.5em,
fill=hidden-orange!60, text width=20em,  text=black,align=left,font=\scriptsize,
inner xsep=2pt,
inner ysep=4pt,
]

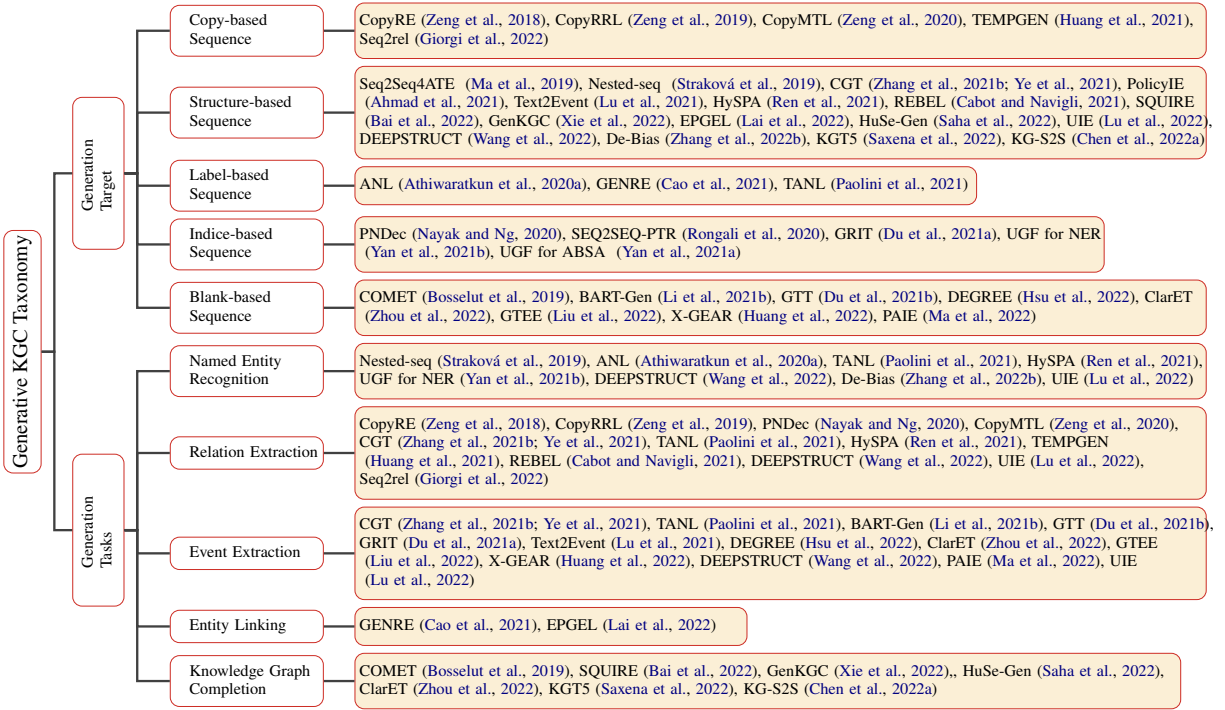
\begin{figure*}[tp]
  \centering
  \resizebox{\textwidth}{!}{
\begin{forest}
  forked edges,
  for tree={
  grow=east,
  reversed=true,
  anchor=base west,
  parent anchor=east,
  child anchor=west,
  base=left,
  font=\small,
  rectangle,
  draw=hiddendraw,
  rounded corners,align=left,
  minimum width=6em,
    edge+={darkgray, line width=1pt},
s sep=3pt,
inner xsep=2pt,
inner ysep=3pt,
ver/.style={rotate=90, child anchor=north, parent anchor=south, anchor=center},
  },
  where level=1{text width=3em,font=\scriptsize,}{},
  where level=2{text width=4.5em,font=\scriptsize,}{},
  [Generative KGC Taxonomy, ver
  [Generation\\Target , ver 
        [Copy-based \\ Sequence  
            [ CopyRE~\citep{DBLP:conf/acl/LiuZZHZ18}{,}
            CopyRRL~\citep{DBLP:conf/emnlp/ZengHZLLZ19}{,}
            CopyMTL~\citep{DBLP:conf/aaai/ZengZL20}{,}
            TEMPGEN~\citep{DBLP:conf/emnlp/HuangTP21}{,}\\
            Seq2rel~\citep{DBLP:conf/bionlp/GiorgiBW22}
            ,leaf,text width=33em]
        ]
        [Structure-based\\Sequence 
            [
            Seq2Seq4ATE ~\citep{DBLP:conf/acl/MaLWXW19}{,}
            Nested-seq  ~\citep{DBLP:conf/acl/StrakovaSH19}{,}
            CGT~\citep{DBLP:journals/taslp/ZhangYDTCHHC21,DBLP:conf/aaai/YeZDCTHC21}{,}
            PolicyIE\\~\citep{DBLP:conf/acl/AhmadCLN0C20}{,}
            Text2Event~\citep{DBLP:conf/acl/0001LXHTL0LC20}{,}
            HySPA~\citep{DBLP:conf/acl/RenSJH21}{,}        
            REBEL~\citep{DBLP:conf/emnlp/CabotN21}{,}
            SQUIRE\\~\citep{DBLP:journals/corr/abs-2201-06206}{,}
            GenKGC~\citep{DBLP:journals/corr/abs-2202-02113}{,}
            EPGEL~\citep{DBLP:conf/acl/LaiJZ22}{,}
            HuSe-Gen~\citep{DBLP:conf/acl/SahaYB22}{,}
            UIE~\citep{DBLP:conf/acl/0001LDXLHSW22}{,}\\
            DEEPSTRUCT~\citep{DBLP:conf/acl/WangLCH0S22}{,}
            De-Bias~\citep{DBLP:conf/acl/Zhang0TW022}{,}
            KGT5~\citep{DBLP:conf/acl/SaxenaKG22}{,}
            KG-S2S~\citep{DBLP:conf/coling/ChenWLL22},
            leaf,text width=33em]
        ]
        [Label-based\\Sequence
            [ 
            ANL~\citep{DBLP:conf/emnlp/AthiwaratkunSKX20}{,}
            GENRE~\citep{DBLP:conf/iclr/CaoI0P21}{,}
            TANL~\citep{DBLP:conf/iclr/PaoliniAKMAASXS21},
            ,leaf,text width=24em]
        ]
        [Indice-based\\Sequence
            [
            PNDec~\citep{DBLP:conf/aaai/NayakN20}{,}
            SEQ2SEQ-PTR~\citep{DBLP:conf/www/RongaliSMH20}{,}
            GRIT~\citep{DBLP:conf/eacl/DuRC21}{,}
            UGF for NER\\~\citep{DBLP:conf/acl/YanGDGZQ20b}{,}
            UGF for ABSA ~\citep{DBLP:conf/acl/YanDJQ020}
            ,leaf,text width=29em]
        ]
        [Blank-based\\Sequence
            [ 
            COMET~\citep{DBLP:conf/acl/BosselutRSMCC19}{,}
            BART-Gen~\citep{DBLP:conf/naacl/LiJH21}{,}
            GTT~\citep{DBLP:conf/naacl/DuRC21}{,}
            DEGREE~\citep{DBLP:journals/corr/abs-2108-12724}{,}
            ClarET\\~\citep{DBLP:conf/acl/ZhouSGLJ22}{,}
            GTEE~\citep{DBLP:conf/acl/LiuHSW22}{,}
            X-GEAR~\citep{DBLP:conf/acl/HuangHNCP22}{,}
            PAIE~\citep{DBLP:conf/acl/MaW0LCWS22}
            ,leaf,text width=33em]
        ]
    ]
    [Generation\\Tasks , ver
    [Named Entity \\ Recognition
    [Nested-seq~\citep{DBLP:conf/acl/StrakovaSH19}{,}
    ANL~\citep{DBLP:conf/emnlp/AthiwaratkunSKX20}{,}
    TANL~\citep{DBLP:conf/iclr/PaoliniAKMAASXS21}{,}
    HySPA~\citep{DBLP:conf/acl/RenSJH21}{,} \\
    UGF for NER~\citep{DBLP:conf/acl/YanGDGZQ20b}{,}
    DEEPSTRUCT~\citep{DBLP:conf/acl/WangLCH0S22}{,}
    De-Bias~\citep{DBLP:conf/acl/Zhang0TW022}{,}
    UIE~\citep{DBLP:conf/acl/0001LDXLHSW22}
    ,leaf,text width=33em
    ]
    ]
    [Relation Extraction
    [CopyRE~\citep{DBLP:conf/acl/LiuZZHZ18}{,}
    CopyRRL~\citep{DBLP:conf/emnlp/ZengHZLLZ19}{,}
    PNDec~\citep{DBLP:conf/aaai/NayakN20}{,}
    CopyMTL~\citep{DBLP:conf/aaai/ZengZL20}{,}\\
    CGT~\citep{DBLP:journals/taslp/ZhangYDTCHHC21,DBLP:conf/aaai/YeZDCTHC21}{,}
    TANL~\citep{DBLP:conf/iclr/PaoliniAKMAASXS21}{,}
    HySPA~\citep{DBLP:conf/acl/RenSJH21}{,} 
    TEMPGEN\\~\citep{DBLP:conf/emnlp/HuangTP21}{,}
    REBEL~\citep{DBLP:conf/emnlp/CabotN21}{,}
    DEEPSTRUCT~\citep{DBLP:conf/acl/WangLCH0S22}{,}
    UIE~\citep{DBLP:conf/acl/0001LDXLHSW22}{,}\\
    Seq2rel~\citep{DBLP:conf/bionlp/GiorgiBW22}
    ,leaf,text width=33em
    ]
    ]
    [Event Extraction
    [CGT~\citep{DBLP:journals/taslp/ZhangYDTCHHC21,DBLP:conf/aaai/YeZDCTHC21}{,}
    TANL~\citep{DBLP:conf/iclr/PaoliniAKMAASXS21}{,}
    BART-Gen~\citep{DBLP:conf/naacl/LiJH21}{,}
    GTT~\citep{DBLP:conf/naacl/DuRC21}{,} \\
    GRIT~\citep{DBLP:conf/eacl/DuRC21}{,}
    Text2Event~\citep{DBLP:conf/acl/0001LXHTL0LC20}{,}
    DEGREE~\citep{DBLP:journals/corr/abs-2108-12724}{,}
    ClarET~\citep{DBLP:conf/acl/ZhouSGLJ22}{,}
    GTEE\\~\citep{DBLP:conf/acl/LiuHSW22}{,}
    X-GEAR~\citep{DBLP:conf/acl/HuangHNCP22}{,}
    DEEPSTRUCT~\citep{DBLP:conf/acl/WangLCH0S22}{,}
    PAIE~\citep{DBLP:conf/acl/MaW0LCWS22}{,}
    UIE\\~\citep{DBLP:conf/acl/0001LDXLHSW22}
    ,leaf,text width=33em
    ]
    ]
    [Entity Linking
    [GENRE~\citep{DBLP:conf/iclr/CaoI0P21}{,}
    EPGEL~\citep{DBLP:conf/acl/LaiJZ22}
    ,leaf,text width=15em
    ]
    ]
    [Knowledge Graph\\ Completion
    [
    COMET~\citep{DBLP:conf/acl/BosselutRSMCC19}{,}
    SQUIRE~\citep{DBLP:journals/corr/abs-2201-06206}{,}
    GenKGC~\citep{DBLP:journals/corr/abs-2202-02113}{,}{,}
    HuSe-Gen~\citep{DBLP:conf/acl/SahaYB22}{,}\\
    ClarET~\citep{DBLP:conf/acl/ZhouSGLJ22}{,}
    KGT5~\citep{DBLP:conf/acl/SaxenaKG22}{,}
    KG-S2S~\citep{DBLP:conf/coling/ChenWLL22}
    ,leaf,text width=32em
    ]
    ]
    ]
]
\end{forest}
}
\caption{Taxonomy of Generative Knowledge Graph Construction.}
\label{taxonomy_of_generation}
\end{figure*}

\subsection{Discrimination and Generation Methodologies}

In this section, we introduce the background of discrimination and generation methodologies for KGC.
The goal of the discrimination model is to predict the possible label based on the characteristics of the input sentence. 
As shown in Figure~\ref{fig:intro}, given annotated sentence $x$ and a set of potentially overlapping triples $t_{j}=\{(s, r, o)\}$ in $x$, we aim to maximize the data likelihood during the training process:
\begin{align}
p_{cls}(t|x)=\prod_{(s, r, o) \in t_{j}} p\left((s, r, o) \mid x_{j}\right)
\end{align}

Another method of discrimination is to output tags using sequential tagging for each position $i$ \citep{DBLP:conf/acl/ZhengWBHZX17,DBLP:conf/aaai/DaiXLDSW19,DBLP:conf/coling/YuZDYZC20,DBLP:conf/coling/LiWZZYC20,DBLP:conf/naacl/LiuFTCZHG21}.
As shown in Figure~\ref{fig:intro}, for an n-word sentence $x$, $n$ different tag sequences are annotated based on "BIESO" (Begin, Inside, End, Single, Outside) notation schema.
The size of a set of pre-defined relations is $|R|$, and the related role orders are represented by "1" and "2". 
During the training model, we maximize the log-likelihood of the target tag sequence using the hidden vector $h_i$ at each position $i$:
\begin{align}
p_{tag}(y \mid x)= \frac{\exp (h_i, y_i)}{\sum_{\mathbf{y}^{\prime} \in R} \exp \left(\exp (h_i, y_i^{\prime})\right)}
\end{align}

For the generation model, if $x$ is the input sentence and $y$ the result of linearized triplets, the target for the generation model is to autoregressively generate $y$ given $x$:
\begin{align}p_{gen}(y \mid x)=\prod_{i=1}^{\operatorname{len}(y)} p_{gen}\left(y_{i} \mid y_{<i}, x\right)
\end{align}

By fine-tuning seq2seq model (e.g.  MASS \citep{DBLP:conf/icml/SongTQLL19} , T5 \citep{DBLP:journals/jmlr/RaffelSRLNMZLL20} , and BART \citep{DBLP:conf/acl/LewisLGGMLSZ20} )  on such a task, using the cross-entropy loss, we can maximize the log-likelihood of the generated  linearized triplets.

\subsection{Advantages of the Generation Methods}

While the previous discriminative methods \citep{DBLP:conf/acl/WeiSWTC20,DBLP:journals/corr/abs-2203-05412} extracts relational triples from unstructured text according to a pre-defined schema to efficiently construct large-scale knowledge graphs, these elaborate models focus on solving a specific task of KGC, such as predicting relation and event information from a segment of input text which often requires multiple models to process.
The idea of formulating KGC tasks as sequence-to-sequence problems \citep{DBLP:conf/acl/0001LDXLHSW22} will be of great benefit to develop a universal architecture to solve different tasks, which can be free from the constraints of dedicated architectures, isolated models, and specialized knowledge sources. 
In addition, generative models can be pre-trained in multiple downstream tasks by structurally consistent linearization of the text, which facilitates the transition from traditional understanding to structured understanding and increases knowledge sharing \citep{DBLP:conf/acl/WangLCH0S22}.
In contexts with nested labels in NER \cite{DBLP:conf/acl/StrakovaSH19}, the proposed generative method implicitly models the structure between named entities, thus avoiding the complex multi-label mapping.
Extracting overlapping triples in RE is also difficult to handle for traditional discriminative models, \citet{DBLP:conf/acl/LiuZZHZ18} introduce a fresh perspective to revisit the RE task with a general generative framework that addresses the problem by end-to-end model.
In short, new directions can be explored for some hard-to-solve problems through paradigm shifts.

Note that the discriminative and generative methods are not simply superior or inferior due to the proliferation of related studies.
The aim of this paper is to summarize the characteristics of different generative paradigms in KGC tasks and provide a promising perspective for future research.

\section{Taxonomy of Generative Knowledge Graph Construction} 
\label{tax}

In this paper, we mainly consider the following five paradigms that are widely used in KGC tasks based on generation target, i.e. \emph{copy-based Sequence}, \emph{structure-linearized Sequence}, \emph{label-augmented Sequence}, \emph{indice-based Sequence}, and \emph{blank-based Sequence}.
As shown in Figure~\ref{fig:tongji}, these paradigms have demonstrated strong dominance in many mainstream KGC tasks.
In the following sections, we introduce each paradigm as shown in Figure~\ref{taxonomy_of_generation}.

\subsection{Copy-based Sequence}
\label{Copy-based}
\begin{figure}
    \centering
    \resizebox{.46\textwidth}{!}{
    \includegraphics{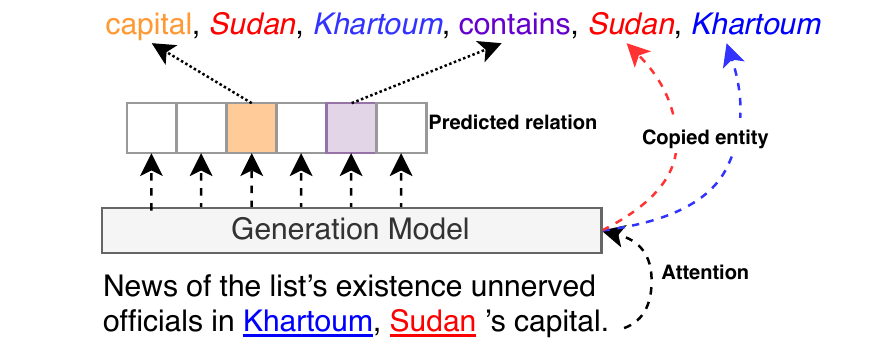}}
    \caption{Copy-based Sequence.}
    \label{fig:copy}
\end{figure}

This paradigm refers to developing more robust models to copy the corresponding token (entity) directly from the input sentence during the generation process. 
\citet{DBLP:conf/acl/LiuZZHZ18} designs an end-to-end model based on a copy mechanism to solve the triple overlapping problem.
As shown in Figure~\ref{fig:copy}, the model copies the head entity from the input sentence and then the tail entity.
Similarly, relations are generated from target vocabulary, which is restricted to the set of special relation tokens.
This paradigm avoids models generating ambiguous or hallucinative entities.
In order to identify a reasonable triple extraction order, \citet{DBLP:conf/emnlp/ZengHZLLZ19} converts the triplet generation process into a reinforcement learning process, enabling the copy mechanism to follow an efficient generative order.
Since the entity copy mechanism relies on unnatural masks to distinguish between head and tail entities, \citet{DBLP:conf/aaai/ZengZL20} 
maps the head and tail entities to fused feature space for entity replication by an additional nonlinear layer, which strengthens the stability of the mechanism.
For document-level extraction, \citet{DBLP:conf/emnlp/HuangTP21} proposes a TOP-k copy mechanism to alleviate the computational complexity of entity pairs.

\subsection{Structure-linearized Sequence}
\label{Structure}
\begin{figure}[t]
    \centering
    \resizebox{.48\textwidth}{!}{
    \includegraphics{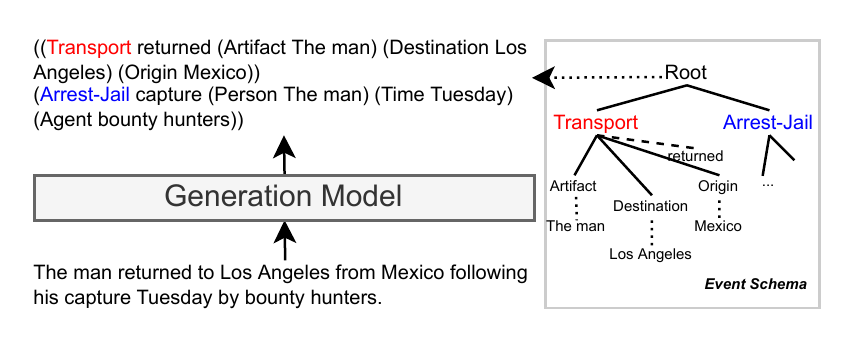}}
    \caption{Structure-linearized Sequence.}
    \label{fig:structure}
\end{figure}

This paradigm refers to utilizing structural knowledge and label semantics, making it prone to handling a unified output format.
\citet{DBLP:conf/acl/0001LXHTL0LC20} proposes an end-to-end event extraction model based on T5, where the output is a linearization of the extracted knowledge structure as shown in Figure~\ref{fig:structure}.
In order to avoid introducing noise, it utilizes the event schema to constrain decoding space, ensuring the output text is semantically and structurally legitimate.
\citet{DBLP:conf/acl/LouLDZC20} reformulates event detection as a Seq2Seq task and proposes a Multi-Layer Bidirectional Network (MLBiNet) to capture the document-level association of events and semantic information simultaneously.
Besides, \citet{DBLP:journals/taslp/ZhangYDTCHHC21,DBLP:conf/aaai/YeZDCTHC21} introduce a contrastive learning framework with a batch dynamic attention masking mechanism to overcome the contradiction in meaning that generative architectures may 
produce unreliable sequences \citep{DBLP:journals/corr/abs-2003-08612}. 
Similarly, \citet{DBLP:conf/emnlp/CabotN21} employs a simple triplet decomposition method for the relation extraction task, which is flexible and can be adapted to unified domains or longer documents.

In the nested NER task, \citet{DBLP:conf/acl/StrakovaSH19}  proposes a flattened encoding algorithm, which outputs multiple NE tags following the BILOU scheme.
The multi-label of a word is a concatenation of all intersecting tags from highest priority to lowest priority.
Similarly, \citet{DBLP:conf/acl/Zhang0TW022} eliminates the incorrect biases in the generation process according to the theory of backdoor adjustment.
In EL task, \citet{DBLP:conf/iclr/CaoI0P21}  proposes Generative ENtity REtrieval (GENRE)  in an autoregressive fashion conditioned on the context, which captures fine-grained interactions between context and entity name.
Moreover, \citet{DBLP:conf/acl/WangLCH0S22,DBLP:conf/acl/0001LDXLHSW22} extends the domain to structural heterogeneous information extraction by proposing a unified task-agnostic generation framework.

\subsection{Label-augmented Sequence} 
~\label{Label-augmented}

\begin{figure}[tb]
    \centering
    \resizebox{.48\textwidth}{!}{
    \includegraphics{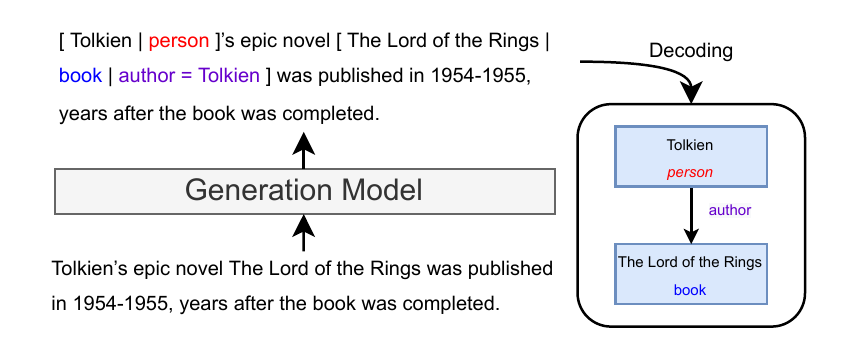}}
    \caption{Label-augmented Sequence.}
    \label{fig:label}
\end{figure}

This paradigm refers to utilizing the extra markers to indicate specific entities or relationships.
As shown in Figure~\ref{fig:label}, \citet{athiwaratkun2020augmented} investigates the label-augmented paradigm for various structure prediction tasks. 
The output sequence copies all words in the input sentence, as it helps to reduce ambiguity.
In addition, this paradigm uses square brackets or other identifiers to specify the tagging sequence for the entity of interest. 
The relevant labels are separated by the separator "$|$" within the enclosed brackets. 
Meanwhile, the labeled words are described with natural words so that the potential knowledge of the pre-trained model can be leveraged \citep{DBLP:conf/iclr/PaoliniAKMAASXS21}. 
Similarly, \citet{DBLP:conf/emnlp/AthiwaratkunSKX20} naturally combines tag semantics and shares knowledge across multiple sequence labeling tasks. 
To retrieve entities by generating their unique names, \citet{DBLP:conf/iclr/CaoI0P21} extends the autoregressive framework to capture the relations between context and entity name by effectively cross-encoding both.
Since the length of the gold decoder targets is often longer than the corresponding input length, this paradigm is unsuitable for document-level tasks because a great portion of the gold labels will be skipped.

\subsection{Indice-based Sequence}
\label{indice}
~\label{Indice-based}

\begin{figure}[tb]
    \centering
    \resizebox{.48\textwidth}{!}{
    \includegraphics{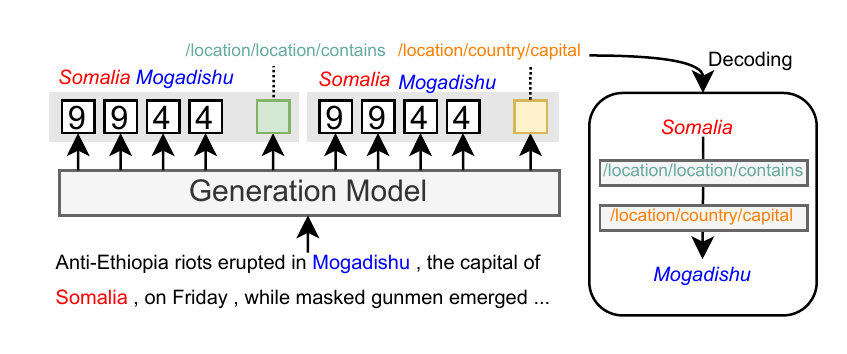}}
    \caption{Indice-based Sequence.}
    \label{fig:indice}
\end{figure}

This paradigm generates the indices of the words in the input text of interest directly and encodes class labels as label indices.
As the output is strictly restricted, it will not generate indices that corresponding entities do not exist in the input text, except for relation labels. 
\citet{DBLP:conf/aaai/NayakN20} apply the method to the relation extraction task, enabling the decoder to find all overlapping tuples with full entity names of different lengths.
As shown in Figure~\ref{fig:indice}, given the input sequence $x$, the output sequence $y$ is generated via the indices: $y = [b_{1},e_{1},t_1,\ldots,b_{i},e_{i},t_i,\ldots, b_{k},e_{k},t_k]$ where $b_i$ and $e_i$ indicates the begin and end indices of a entity tuple, 
$t_i$ is the index of the entity type, and $k$ is the number of entity tuples. 
The hidden vector is computed at decoding time by the pointer network ~\citep{vinyals2015pointer} to get the representation of the tuple indices.
Besides, \citet{DBLP:conf/acl/YanGDGZQ20b} explores the idea of generating indices for NER, which can be applied to different settings such as flat, nested, and discontinuous NER. 
In addition, \citet{DBLP:conf/eacl/DuRC21} applies the method to a role-filler entity extraction task by implicitly capturing noun phrase coreference structure.

\subsection{Blank-based Sequence}
~\label{Blank-based}

\begin{figure}[t]
    \centering
    \resizebox{.48\textwidth}{!}{
    \includegraphics{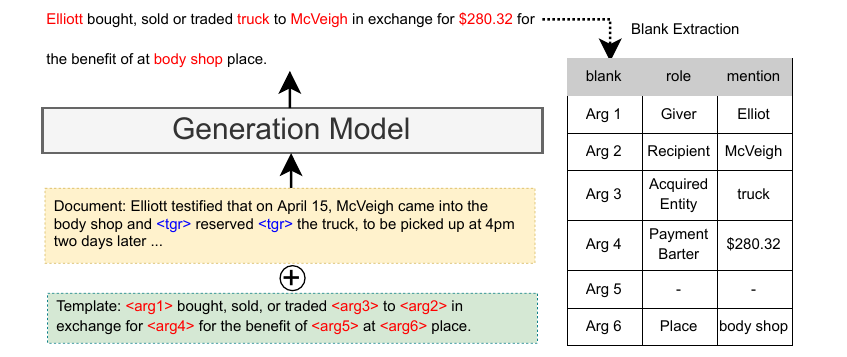}}
    \caption{Blank-based Sequence.}
    \label{fig:blank}
\end{figure}

This paradigm refers to utilizing templates to define the appropriate order and relationship for the generated spans.
\citet{DBLP:conf/naacl/DuRC21} explores a blank-based form for event extraction tasks which includes special tokens representing event information such as event types. 
\citet{DBLP:conf/naacl/LiJH21} frames document-level event argument extraction as conditional generation given a template and introduces the new document-level informative to aid the generation process. 
As shown in Figure~\ref{fig:blank},  the template refers to a text describing an event type, which adds blank argument role placeholders.
The output sequences are sentences where the blank placeholders are replaced by specific event arguments. 
Besides, \citet{DBLP:journals/corr/abs-2108-12724} focuses on low-resource event extraction and proposes a data-efficient model called DEGREE, which utilizes label semantic information.
\citet{DBLP:conf/acl/HuangHNCP22} designs a language-agnostic template to represent the event argument structures, which facilitate the cross-lingual transfer.
Instead of conventional heuristic threshold tuning, \citet{DBLP:conf/acl/MaW0LCWS22} proposes an effective yet efficient model PAIE for extracting multiple arguments with the same role. 

\subsection{Comparison and Discussion}
 
\begin{table*}[t!]
\centering
\label{table:entity}
\resizebox{\textwidth}{!}{
    \begin{tabular}{ l l l cccc }
    \toprule
    \multirow{2}{*}{\textbf{Taxonomy}} & \multirow{2}{*}{\textbf{Generative Strategy}} & \multirow{2}{*}{\textbf{Representative Model}} & \multicolumn{4}{c}{\textbf{Evaluation Scope}} \\ 
    \cline{4-7}
     & & & SU$\uparrow$ & SS$\downarrow$  & AS$\downarrow$ & TS$\downarrow$  \\ 
     \toprule 
   \multirow{2}{*}{Copy-based  (\S~\ref{Copy-based})} 
        & Directly copy entity & CopyRE~\citep{DBLP:conf/acl/LiuZZHZ18}
        &  L  & L    & M  & L      \\
        & Restricted target vocabulary &Seq2rel~\citep{DBLP:conf/bionlp/GiorgiBW22}
        &  L  & L    & H  & L      \\
    \midrule
   \multirow{8}{*}{Structure-based (\S~\ref{Structure})} 
           & Per-token tag encoding & Nested-seq~\citep{DBLP:conf/acl/StrakovaSH19}
        &  L  &  L  & H  & L   \\
      & Faithful contrastive learning & CGT~\citep{DBLP:journals/taslp/ZhangYDTCHHC21}
        &  M  & M     & H  & L    \\ 
        & Prefix tree constraint decoding & TEXT2EVENT~\citep{DBLP:conf/acl/0001LXHTL0LC20}
        & M  &  M     &  H  & L  \\
         &  Triplet linearization & REBEL~\citep{DBLP:conf/emnlp/CabotN21}
        &  M   & H  & M   & L     \\
         & Entity-aware hierarchical decoding & GenKGC~\citep{DBLP:journals/corr/abs-2202-02113}
        &  M   & L  & M   & L     \\
          &  Unified structure generation & UIE~\citep{DBLP:conf/acl/0001LDXLHSW22}
        & M   &  H  & H  & L  \\
        &  Reformulating triple prediction & DEEPSTRUCT~\citep{DBLP:conf/acl/WangLCH0S22}
        & M   &  H  & H  & L  \\
        & Query Verbalization & KGT5~\citep{DBLP:conf/acl/SaxenaKG22}
        & M   &  H  & M  & L  \\
        
    \midrule
   \multirow{1}{*}{Label-based (\S~\ref{Label-augmented})} 
        & Augmented natural language & TANL~\citep{DBLP:conf/iclr/PaoliniAKMAASXS21}
        & M   & H  & H  & L     \\
    \midrule 
   \multirow{2}{*}{Indice-based (\S~\ref{Indice-based})} 
        & Pointer mechanism &  PNDec~\citep{DBLP:conf/aaai/NayakN20}
        & L  &  L  & M  & L   \\
        & Pointer selection &  GRIT~\citep{DBLP:conf/eacl/DuRC21}
        & M  &  L  & M  & L   \\
   \midrule 
   \multirow{3}{*}{Blank-based (\S~\ref{Blank-based})} 
        & Template filling as generation &  GTT~\citep{DBLP:conf/naacl/DuRC21}
        & H  & H    & H   & H  \\
    & Prompt semantic guidance &  DEGREE~\citep{DBLP:journals/corr/abs-2108-12724}
        & H  & H    & H   & H  \\
    & Language-agnostic template &  X-GEAR~\citep{DBLP:conf/acl/HuangHNCP22}
        & H  & M    & H   & H  \\
    \bottomrule
    \end{tabular}
}
\caption{
    Comparison of generation methods from different evaluation scopes. "SU" indicates semantic utilization, "SS" indicates search space, 
   "AS" indicates application scope, and "TS" indicates template cost.
   We divide the degree into three grades$:$ L (low), M (middle), and H (high), and the $\uparrow$ indicates that the higher grade performance is better while the $\downarrow$ is the opposite.
   }
 \label{fig:sum}
\end{table*}

Recently, the literature on generative KGC has been growing rapidly.
A unifying theme across many of these methods is that of end-to-end architecture or the idea that the knowledge extraction can be redefined as \emph{text sequence to structure generation} task.
Generative models can decode and control extraction targets on demand for different specific tasks, scenarios, and settings (i.e., different schema).
However, due to the different forms of specific KGC tasks, there is still some disagreement in the utilization of the generation paradigms. 

As shown in Table~\ref{fig:sum}, we make a comprehensive comparison among the paradigms mentioned above via rating based on different evaluation scopes: 
1) \textbf{Semantic utilization} refers to the degree to which the model leverages the semantics of the labels. 
In principle, we believe that the closer the output form is to natural language, the smaller the gap between the generative model and the training task.
We observe that the blank-based paradigm has a clear advantage in this scope, which uses manually constructed templates to make the output close to natural language fluency. 
2) \textbf{Search space} refers to the vocabulary space searched by the decoder. 
Due to the application of the constraint decoding mechanism, some structure-based methods can be reduced to the same decoding space as the copy-based methods. 
In addition, the indice-based paradigm uses a pointer mechanism that constrains the output space to the length of the input sequence.
3) \textbf{Application scope} refers to the range of KGC tasks that can be applied.
We believe that architectures with the ability to organize information more flexibly have excellent cross-task migration capabilities such as structure-based, label-based and blank-based paradigms.
4) \textbf{Template cost} refers to the cost of constructing the input and golden output text.
We observe that most paradigms do not require complex template design and rely only on linear concatenation to meet the task requirement. 
However, the blank-based paradigm requires more labor consumption to make the template conform to the semantic fluency requirement.

Totally in line with recent trends in NLP, a growing number of unified generation strategies require more universal architectures \citep{DBLP:conf/emnlp/DengTLXH21, DBLP:conf/acl/LiTHJHXCYZW21}, as they allow a remarkable degree of output flexibility.
We think that future research should focus on unifying cross-task models and further improving decoding efficiency.
\begin{table*}[t] \centering
\resizebox{0.628\textwidth}{!}{
\begin{tabular}{ll|ccc}
\toprule
\multirow{2}{*}{Type}           & \multirow{2}{*}{Models} & \multicolumn{3}{c}{\textbf{NYT}}  \\
\cline{3-5}
&& P & R & F   \\
\hline
\hline
   \multirow{3}{*}{Discrimination}    
    & CasRel~\citep{DBLP:conf/acl/WeiSWTC20}  &89.7 &89.5 &89.6\\ 
    & TPLinker~\citep{DBLP:conf/coling/WangYZLZS20}  &91.4 &92.6 &92.0 \\
    & OneRel~\citep{DBLP:journals/corr/abs-2203-05412} & 92.8 &  92.9 & 92.8\\
\hline
\hline

   \multirow{3}{*}{Copy-based }   
   &  CopyRE~\citep{DBLP:conf/acl/LiuZZHZ18} & 61.0& 56.6& 58.7\\
       & CopyRRL~\citep{DBLP:conf/emnlp/ZengHZLLZ19}  & 77.9 &67.2 &72.1 \\
    & CopyMTL~\citep{DBLP:conf/aaai/ZengZL20}   &75.7  &68.7  &72.0   \\

\hline
\multirow{4}{*}{Structure-based } 
     & CGT~\citep{DBLP:conf/aaai/YeZDCTHC21} & 94.7 &84.2 &89.1 \\
    & REBEL~\citep{DBLP:conf/emnlp/CabotN21}  &91.5 &92.0 &91.8 \\
     & UIE~\citep{DBLP:conf/acl/0001LDXLHSW22} & - &- &93.5 \\
     & DEEPSTRUCT~\citep{DBLP:conf/acl/WangLCH0S22} & - &- &93.9
     \\
     \hline
\multirow{1}{*}{Label-based} 
    & TANL~\citep{DBLP:conf/iclr/PaoliniAKMAASXS21} &- &- & 90.8 \\
\hline  
\multirow{1}{*}{Indice-based} 
    & PNDec~\citep{DBLP:conf/aaai/NayakN20} & 89.3& 78.8& 83.8 \\
\hline  
\multirow{2}{*}{Others*} 
    & SPN~\citep{DBLP:journals/corr/abs-2011-01675} &93.3 &91.7 &92.5 \\
    & Seq2UMTree~\citep{DBLP:conf/emnlp/ZhangLFJZCKK20} & 79.1 &75.1 & 77.1\\
\bottomrule 
\end{tabular}
}
\caption{Main results of NYT dataset. The top section refers to the discrimination models, and the bottom section indicates generation models. "*" refers to the non-autoregressive models. }
\label{nyt}
\end{table*}

\begin{table*}[!t] \centering
\resizebox{0.74\textwidth}{!}{
\begin{tabular}{l|l|cc|cc}
\toprule
\multirow{2}{*}{Type}           & \multirow{2}{*}{Models}                                & \multicolumn{2}{c|}{Trigger} & \multicolumn{2}{c}{Argument} \\ \cline{3-6} 
                                &                                                        & Id            & Cl           & Id            & Cl           \\ 
\hline
\multirow{6}{*}{Discrimination} & JMEE \cite{DBLP:conf/emnlp/LiuLH18} & 75.9              & 73.7             &  68.4             & 60.3       \\
& DYGIE++ \citep{DBLP:conf/emnlp/WaddenWLH19} &- &69.7 &53.0 & 48.8\\ 
& OneIE \citep{DBLP:conf/acl/LinJHW20} & 78.6 & 75.2 & 60.7 &58.6 \\ 
    & QAEE \citep{DBLP:conf/emnlp/DuC20} &75.8 &72.4 &55.3 &53.3 \\
    & MQAEE  \citep{DBLP:conf/emnlp/LiPCWPLZ20}  &74.5 &71.7 & 55.2 &53.4 \\
    & RCEE \citep{DBLP:conf/emnlp/LiuCLBL20} &- &74.9 &- &63.6 \\ 

\hline
\hline
\multirow{3}{*}{Structure-based} 
     & TEXT2EVENT \citep{DBLP:conf/acl/0001LXHTL0LC20} & - & 71.9 & - & 53.8 \\ 
    & UIE \citep{DBLP:conf/acl/0001LDXLHSW22} & - & 73.4 & - & 54.8 \\
    & DEEPSTRUCT \citep{DBLP:conf/acl/WangLCH0S22} & 73.5 & 69.8 &  59.4 & 56.2 \\
\hline  
\multirow{1}{*}{Label-based} 
    & TANL \cite{DBLP:conf/iclr/PaoliniAKMAASXS21} &72.9 &68.4 &50.1 &47.6 \\
\hline  
\multirow{4}{*}{Blank-based} 
    & BART-Gen \citep{DBLP:conf/naacl/DuRC21} &74.4 &71.1 &55.2 &53.7 \\ 
    & DEGREE \citep{DBLP:journals/corr/abs-2108-12724} &- &73.3 &- &55.8\\
     & GTEE \citep{DBLP:conf/acl/LiuHSW22} &- &72.6 &- &55.8\\ 
     &PAIE \citep{DBLP:conf/acl/MaW0LCWS22} &- &- &75.7$^{*}$ & 72.7$^{*}$\\ 
     \bottomrule

\end{tabular}
}

\caption{F1  results (\%) of ACE-2005. The top section refers to the discrimination models, and the bottom section indicates the generation models. Id is Identification, and Cl is Classification.  "*" refers to experiments only in argument extraction tasks with the golden trigger.} 
\label{ace}
\end{table*}
 
\section{Analysis} 
\label{analysis}

\subsection{Theoretical Insight} 

This section provides theoretical insight into optimization and inference for generative KGC.
For optimization, NLG  are normally modeled by parameterized probabilistic models $p_{gen}$ over text strings $\yy = \langle y_1, y_2, \dots \rangle$ decomposed by words $y_t$:
\begin{align}
p_{gen}(y \mid x)=\prod_{i=1}^{\operatorname{len}(y)} p_{gen}\left(y_{i} \mid y_{<i}, x\right)
\end{align} 
where $y$  consists of all possible strings that can be constructed from words in the model's vocabulary $\vocab$. 
Note that the output $y$ can take on a variety of forms depending on the task, e.g., entities, relational triples, or an event structure.
Usually, the model will limit the target set by pre-defined schema as  $\calY_\mathcal{T} \subset \calY$.
The optimization procedure will be taken to estimate the parameters  with log-likelihood  maximization as follows:
\begin{align}\label{eq:nll}
        L(\vtheta;\mathcal{T}) = -\sum_{\yy \in \mathcal{T}}\log \model(\yy)
\end{align} 
where  $\vtheta$ are the model parameters.
Notably, with small output space (e.g., methods with the indice-based sequence in \S \ref{indice}), the model can converge faster. 
However, the model with a small output space may fail to utilize rich semantic information from labels or text (like models in \S \ref{Blank-based}).
In short, the design of output space is vital for generative KGC, and it is necessary to balance parametric optimization as well as semantic utilization.

For inference, we argue that sequence decoding in the generation is an essential procedure for generative KGC. 
Given the probabilistic nature of $\model$, the decoding process will select words that maximize the probability of the resulting string.
Vanilla decoding solutions such as beam search or greedy have been investigated in generative KGC. 
On the one hand, knowledge-guided (or schema-guided) decoding has become the mainstay for many generative KGC tasks. 
For example, \citet{DBLP:conf/acl/0001LXHTL0LC20} proposes Text2Event in which words are decoded through a prefix tree based on pre-defined  schema.  
On the other hand, non-autoregressive parallel decoding has also been leveraged for generative KGC. 
\citet{DBLP:conf/emnlp/SuiW000B21} formulates end-to-end knowledge base population as a direct set generation problem, avoiding considering the order of multiple facts.
Note that the decoding mechanism plays a vital role in inference speed and quality.
We argue that it is necessary to develop sophisticated, efficient decoding strategies (e.g., with guidance from KG) for generative KGC.

\subsection{Empirical Analysis}   

To investigate the effect of different generation methods, we conduct an analysis of the experimental results of existing generative KGC work. 
Due to space limitations of the article, we only select two representative tasks of entity/relation extraction and event extraction with NYT and ACE datasets\footnote{Results are taken from existing papers.}.
Table \ref{nyt} shows the performance of discrimination models and generative models on the NYT datasets.
We can observe that:
1) Structure-based and label-based methods both achieve similar extraction performance compared with all discrimination models on NYT datasets. 
We believe this is because they can better utilize label semantics and structural knowledge than other generation methods.
2) Although the discrimination methods obtain good performance, the performance of the generation methods has been improved more vastly in recent years, so we have reason to believe that they will have greater application scope in the near future.
In addition, we also show the performance of the non-autoregressive method on two datasets, and we discuss the promising value of this method in \S~\ref{future work}.
We observe that parallel generation of the unordered triple set can obtain comparable performance with advanced discriminative models, noting that non-autoregressive methods have better decoding efficiency and training efficiency.

From Table \ref{ace}, we observe that generation methods can obtain comparable performance compared with discrimination models on event extraction tasks.
Since the framework of event extraction has a hierarchical structure (i.e., it is usually decomposed into two subtasks: trigger extraction and argument extraction), 
structure-based methods have a supervised learning framework for the sequence-to-structure generation, while schema constraints guarantee structural and semantic legitimacy.
In addition, owing to the complete template design of the Blank-based approach, PLMs can understand complex task knowledge, structural knowledge of the extraction framework, and label semantics in a natural language manner.

\section{Future Directions}
\label{future work}
Though lots of technical solutions have been proposed for generative KGC as surveyed, there remain some potential directions:

\textbf{Generation Architecture.}
Most of the recent generative KGC frameworks face serious homogenization with Transformer.
For enhancing interpretability, we argue that neuro-symbolic models (i.e., a reasoning system that integrates neural and symbolic) \citep{DBLP:journals/aiopen/ZhangCZKD21,DBLP:phd/basesearch/Galassi21,DBLP:journals/aepia/NegroP22} can be designed for generative KGC.
In addition, some cutting-edge technologies such as spiking neural network \citep{DBLP:journals/nn/TavanaeiGKMM19}, dynamic neural networks \citep{DBLP:journals/corr/abs-2202-07101}, ordinary differential equations \citep{DBLP:conf/acl/LiDZJZZXZ0Z22} and diffusion models \citep{DBLP:conf/nips/DhariwalN21} can also provide promising architectures.

\textbf{Generation Quality.}
Considering the target reliability of generation methods, more sophisticated strategies can be leveraged  to control the quality of generative KGC, including:
1) Control code construction \citep{DBLP:journals/corr/abs-1909-05858,DBLP:conf/naacl/DouLHJN21};
2) Decoding strategy such as introducing external feedback \citep{DBLP:conf/acl/ChoiBGHBF18} and generative discriminator \citep{DBLP:conf/emnlp/KrauseGMKJSR21};
3) Loss function design \citep{DBLP:conf/iclr/ChanOPZF21};
4) Prompt design \citep{DBLP:conf/nips/BrownMRSKDNSSAA20,DBLP:conf/acl/Qian0SWC22};
5) Retrieval augmentation \citep{DBLP:journals/corr/abs-2202-01110};
6) Write-then-Edit strategy
\citep{DBLP:conf/iclr/DathathriMLHFMY20};
7) Diffusion process \citep{DBLP:journals/corr/abs-2205-14217,https://doi.org/10.48550/arxiv.2210.08933}.

\textbf{Training Efficiency.}
In practical applications, it is essential to reduce data annotation and training costs.
One idea is to freeze most of the generation model parameters \cite{DBLP:journals/corr/abs-2103-10385,DBLP:conf/acl/LiL20,chen-etal-2022-lightner} or leverage prompt learning \citep{DBLP:conf/www/ChenZXDYTHSC22}.
Another idea is that knowledge decoupling intervention training models can reduce parameter redundancy \citep{DBLP:conf/acl/WangTDWHJCJZ21,DBLP:journals/corr/abs-2112-04426,DBLP:conf/iclr/KhandelwalLJZL20,DBLP:journals/corr/abs-2205-02355,DBLP:journals/corr/abs-2205-14704}.

\textbf{Universal Deployment.}
Inspired by the T5 \citep{DBLP:journals/jmlr/RaffelSRLNMZLL20}, which transforms all NLP tasks into Text-to-Text tasks, generation models can be generalized to the multi-task and multi-modal domain.
Therefore, instead of improvements being prone to be exclusive to a single task, domain, or dataset, we argue that it is beneficial to study the framework to advocate for a unified view of KGC, such as the wonderful work UIE \citep{DBLP:conf/acl/0001LDXLHSW22}. 
Furthermore, it is efficient for real-world deployment when we can provide a single model to support widespread KGC tasks \citep{EMNLP2020_OpenUE}.

\textbf{Inference Speed.}
To be noted, although previous work has treated KGC as end-to-end generative tasks, they are still limited by auto-regressive decoders.
However, the autoregressive decoder generates each token based on previously generated tokens during inference, and this process is not parallelizable. 
Therefore, it is beneficial to develop a fast inference model for generative KGC.
Previously, \citet{DBLP:journals/corr/abs-2011-01675} utilizes the transformer-based non-autoregressive decoder \citep{DBLP:conf/iclr/Gu0XLS18} as a triple set generator that can predict all triples at once.
\citet{DBLP:conf/emnlp/SuiW000B21} also formulates end-to-end  knowledge base population as a direct set generation problem.
\citet{DBLP:conf/emnlp/ZhangLFJZCKK20} proposes a two-dimensional unordered multitree allowing prediction deviations not to aggregate and affect other triples. 
To sum up, the non-autoregressive approach applied to KGC proves to be effective in solving the exposure bias and overfitting problems.
Likewise, the semi-autoregressive decoding \citep{DBLP:conf/emnlp/WangZC18} preserves the autoregressive approach within the block to ensure consistency while improving the tuple output efficiency.
Additionally, adaptive computation \citep{DBLP:conf/acl/SchwartzSSDS20} can accelerate inference by ignoring inner modules, which is faster and more efficient as it does not activate the entire network for knowledge graph construction.


\section{Conclusion and Vision} 

In this paper, we provide an overview of generative KGC with new taxonomy, theoretical insight and empirical analysis, and several research directions.
Note that the generative paradigm for KGC has the potential advantages of unifying different tasks and better utilizing semantic information. 
In the future, we envision a more potent synergy between the methodologies from the NLG and knowledge graph communities.
We hope sophisticated and efficient text generation models to be increasingly contributed to improving the KGC  performance.
On the converse, we expect symbolic structure in KG can have potential guidance for text generation. 

\section{Limitations} 
In this study, we provide a review of generative KGC. 
Due to the page limit, we cannot afford the technical details for models. 
Moreover, we only review the works within five years, mainly from the ACL, EMNLP, NAACL, COLING, AAAI, IJCAI, etc. 
We will continue adding more related works with more detailed analysis. 

\section*{Acknowledgment}
We want to express gratitude to the anonymous reviewers.
This work was supported by the National Natural Science Foundation of China (No.62206246, 91846204 and U19B2027), Zhejiang Provincial Natural Science Foundation of China (No. LGG22F030011), Ningbo Natural Science Foundation (2021J190), and Yongjiang Talent Introduction Programme (2021A-156-G). 
This work was supported by Information Technology Center and State Key Lab of CAD\&CG, ZheJiang University.

\bibliography{anthology,custom}
\bibliographystyle{acl_natbib}

\appendix

\section{Timeline Analysis}
\label{sec:timeline}

As shown in Table~\ref{timeline}, we summarize a number of existing research papers in chronological order in the form of a timeline, which hopefully helps researchers who are new to this topic understand the evolution of the generative KGC paradigms.

\definecolor{myorange}{rgb}{1.0, 0.49, 0.0}	
\definecolor{tlgreen}{rgb}{0.33, 0.68, 0.20}
\definecolor{mypurple}{rgb}{0.60, 0.0,  0.60}

\begin{table*}[b]
\scriptsize
\renewcommand\arraystretch{2}\arrayrulecolor{black}
\captionsetup{singlelinecheck=false, font=black, labelfont=sc, labelsep=quad}
\caption{Timeline of generative KGC. The time for each paper is based on its first arXiv version (if it exists) or estimated submission time. 
Works in \textcolor{red}{red} consider copy-based sequence methods; 
works in \textcolor{blue}{blue} consider structure-linearized sequence methods; 
works in \textcolor{tlgreen}{green} consider label-augmented sequence methods;
works in \textcolor{myorange}{orange} consider
indice-based sequence methods;
works in \textcolor{mypurple}{purple} consider
blank-based sequence methods.}\vskip -1.5ex
\begin{tabular}{
@{\,}r <{\hskip 2pt} 
!{\foo}
>{\raggedright\arraybackslash}p{6.5cm}@{\,}r <{\hskip 2pt} !{\foo} >{\raggedright\arraybackslash}p{5.5cm}}
\toprule
\addlinespace[1.5ex]

2018.06.15 & \textcolor{red}{CopyRE} \citep{DBLP:conf/acl/LiuZZHZ18} 
&2021.09.10 & \textcolor{red}{TEMPGEN} \citep{DBLP:conf/emnlp/HuangTP21}\\

2019.06.12 & \textcolor{mypurple}{COMET} \citep{DBLP:conf/acl/BosselutRSMCC19} &2021.11.07 & \textcolor{blue}{REBEL} \citep{DBLP:conf/emnlp/CabotN21}  \\

2019.07.28 & \textcolor{blue}{Seq2Seq4ATE} \citep{DBLP:conf/acl/MaLWXW19} 
&2022.01.17 & \textcolor{blue}{SQUIRE} \citep{DBLP:journals/corr/abs-2201-06206}\\

2019.08.19 & \textcolor{blue}{Nested-seq} \citep{DBLP:conf/acl/StrakovaSH19}
&2022.02.04 & \textcolor{blue}{GenKGC} \citep{DBLP:journals/corr/abs-2202-02113}\\

2019.11.04 & \textcolor{red}{CopyRRL} \citep{DBLP:conf/emnlp/ZengHZLLZ19}
&2022.02.27 & \textcolor{blue}{EPGEL} \citep{DBLP:conf/acl/LaiJZ22}  
 \\

2019.11.22 & \textcolor{myorange}{PNDec} \citep{DBLP:conf/aaai/NayakN20}  
&2022.04.11 & \textcolor{blue}{HuSe-Gen} \citep{DBLP:conf/acl/SahaYB22} 
 \\

2019.11.24 & \textcolor{red}{CopyMTL} \citep{DBLP:conf/aaai/ZengZL20} 
&2022.05.04 & \textcolor{mypurple}{ClarET} \citep{DBLP:conf/acl/ZhouSGLJ22} 
 \\

2020.01.30 & \textcolor{myorange}{SEQ2SEQ-PTR} \citep{DBLP:conf/www/RongaliSMH20}
&2022.05.12 & \textcolor{mypurple}{GTEE} \citep{DBLP:conf/acl/LiuHSW22} 
 \\

2020.09.14 & \textcolor{blue}{CGT} \citep{DBLP:journals/taslp/ZhangYDTCHHC21}
& 2022.05.15 & \textcolor{mypurple}{X-GEAR} \citep{DBLP:conf/acl/HuangHNCP22}   \\

2020.09.15 & \textcolor{tlgreen}{ANL} \citep{DBLP:conf/emnlp/AthiwaratkunSKX20} 
&2022.05.22 & \textcolor{blue}{DEEPSTRUCT} \citep{DBLP:conf/acl/WangLCH0S22}\\

2020.10.02 & \textcolor{tlgreen}{GENRE} \citep{DBLP:conf/iclr/CaoI0P21}
&2022.05.22 & \textcolor{blue}{De-Bias} \citep{DBLP:conf/acl/Zhang0TW022}\\

2021.01.01 & \textcolor{blue}{PolicyIE} \citep{DBLP:conf/acl/AhmadCLN0C20}
&2022.05.22 & \textcolor{blue}{KGT5} \citep{DBLP:conf/acl/SaxenaKG22} \\

2021.01.14 & \textcolor{tlgreen}{TANL} \citep{DBLP:conf/iclr/PaoliniAKMAASXS21}
&2022.05.22 & \textcolor{mypurple}{PAIE}
\citep{DBLP:conf/acl/MaW0LCWS22} \\

2021.04.13 & \textcolor{mypurple}{BART-Gen} \citep{DBLP:conf/naacl/LiJH21} 
& 2022.05.23 & \textcolor{blue}{UIE} \citep{DBLP:conf/acl/0001LDXLHSW22}\\

2021.04.21 & \textcolor{myorange}{GRIT} \citep{DBLP:conf/eacl/DuRC21}
& 2022.09.15 & \textcolor{red}{Seq2rel}
\citep{DBLP:conf/bionlp/GiorgiBW22}
\\

2021.06.02 & \textcolor{myorange}{UGF for NER} \citep{DBLP:conf/acl/YanGDGZQ20b}
& 2022.09.15 & \textcolor{blue}{KG-S2S}
\citep{DBLP:conf/coling/ChenWLL22}
\\

2021.06.08 & \textcolor{myorange}{UGF for ABSA} \citep{DBLP:conf/acl/YanDJQ020}  \\

2021.06.11 & \textcolor{mypurple}{GTT} \citep{DBLP:conf/naacl/DuRC21}  \\

2021.06.17 & \textcolor{blue}{Text2Event} \citep{DBLP:conf/acl/0001LXHTL0LC20}  \\

2021.06.30 & \textcolor{blue}{HySPA} \citep{DBLP:conf/acl/RenSJH21}  \\

2021.08.29 & \textcolor{mypurple}{DEGREE} \citep{DBLP:journals/corr/abs-2108-12724}\\

\end{tabular} \label{timeline}
\end{table*}

\end{document}